\documentclass[letterpaper, 10 pt, conference]{ieeeconf}  %
\usepackage[font=small, labelfont=bf]{caption}
\usepackage{multicol}
\usepackage[bookmarks=true, pagebackref]{hyperref}
\usepackage{lipsum}
\usepackage{xspace}
\usepackage{todonotes}
\usepackage{amsmath}
\usepackage{amssymb}
\usepackage{booktabs} 
\usepackage{wrapfig}
\usepackage[english]{babel}
\usepackage{here}
\usepackage{caption}
\usepackage{subcaption}
\usepackage{algorithm}
\usepackage{algpseudocode}
\usepackage{float}
\usepackage{multirow}

\newtheorem{hypothesis}{Hypothesis}

\IEEEoverridecommandlockouts                              %

\newcommand{\edit}[1]{{\color{black} #1\xspace}}

\title{\LARGE \bf
Staircase Localization for Autonomous Exploration \\ in Urban Environments

}

\author{Jinrae Kim$^{1}$, Sunggoo Jung$^{2}$, Sung-Kyun Kim$^{2}$, Youdan Kim$^{1}$, Ali-akbar Agha-mohammadi$^{2}$%
\thanks{$^{1}$ Jinrae Kim and Youdan Kim are with the Department of Aerospace Engineering, Seoul National University (SNU),  Institute of Advanced Aerospace Technology, Seoul 08826, Republic of Korea. {\tt\small \{kjl950403,ydkim\}@snu.ac.kr}}%
\thanks{$^{2}$ Sunggoo Jung, Sung-Kyun Kim, and Ali-akbar Agha-mohammadi are with the Mobility and Robotic Systems section at NASA Jet Propulsion Laboratory (JPL), California Institute of Technology, Pasadena 91109, CA, USA {\tt\small \{sunggoo.jung, sung.kim, aliakbar.aghamohammadi\}@jpl.nasa.gov}}%
}
\begin{document}

\maketitle
\thispagestyle{empty}
\pagestyle{empty}

\begin{abstract}
A staircase localization method is proposed for robots to explore urban environments autonomously.
The proposed method employs a modular design in the form of a cascade pipeline consisting of three modules of stair detection, line segment detection, and stair localization modules.
The stair detection module utilizes an object detection algorithm based on deep learning to generate a region of interest (ROI).
From the ROI, line segment features are extracted using a deep line segment detection algorithm.
The extracted line segments are used to localize a staircase in terms of position, orientation, and stair direction.
The stair detection and localization are performed only with a single RGB-D camera. Each component of the proposed pipeline does not need to be designed particularly for staircases, which makes it easy to maintain the whole pipeline and replace each component with state-of-the-art deep learning detection techniques.
\edit{
The results of real-world experiments show that the proposed method can perform accurate stair detection and localization during autonomous exploration for various structured and unstructured upstairs and downstairs with shadows, dirt, and occlusions by artificial and natural objects.
}

\end{abstract}

\section{Introduction}
Staircase is a common feature found in urban environments that can pose a challenge for robots to navigate.
Therefore, the ability to accurately detect and localize staircases is important for a variety of urban applications, including autonomous exploration and aiding visually impaired individuals~\cite{otsu2020supervised,vu2008autonomous,munoz2016depth}.
In the literature, stair detection can refer to different things depending on the context:
i) detecting stairs as objects in a color image~\cite{patil2019deep},
ii) extracting features such as line segments from a color image~\cite{wang2022deep},
or iii) perceptual detection using a laser range finder that follows a set of rules~\cite{mihankhah2009autonomous}.
In this study, the term ``stair detection" refers to the first definition.\\
Stair detection involves identifying the presence and location of stairs in an image or video. This process can be used to draw attention to a specific region of the image, known as a region of interest (ROI). In addition, the stair line segment detection can provide valuable visual features for higher-level tasks in computer vision~\cite{gu2022towards}. Stair localization, on the other hand, involves determining the position of stairs within an image, and may also include the information about the orientation and direction of the stairs (e.g., whether they lead up or down). 

Several methods have been proposed that the visual appearance of a staircase, specifically the presence of parallel lines, can give clues about the pose and orientation of the staircase.
In~\cite{munoz2016depth}, Munoz et al. used traditional computer vision techniques and optical flow to detect parallel lines on a staircase,
and then used a support vector machine to classify the staircase as going up, going down, or not a staircase.
In~\cite{patil2019deep}, Patil et al. used a deep object detection algorithm called ``You Only Look Once" (YOLO) to detect stairs, and then applied traditional computer vision techniques to extract parallel lines within the ROI provided by the object detection algorithm.
\begin{figure}[!t]
    \centering
    \includegraphics[width=1.0\columnwidth]{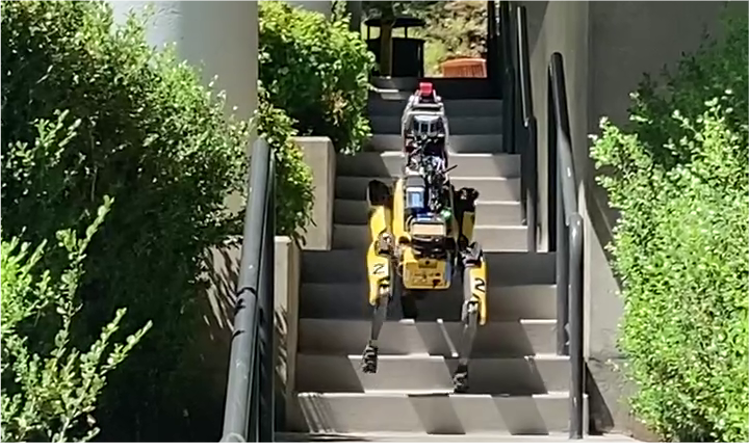}
    \caption{In our autonomous urban exploration system, an RGB-D camera is used to detect and localize stairs.
    A vision-based deep learning approach is employed to detect the stairs and provide an ROI. Within the ROI, we use line segments and corresponding depth data to estimate the pose of the stairs. This allows the robot to perform real-time stair localization from multiple perspectives.}
    \label{fig:banner_fig}
\end{figure}
Lu and Manduchi proposed a stair detection and localization algorithm based on traditional computer vision techniques such as Canny edge detection and Hough transform, combined with stereo vision~\cite{lu2005detection}. These techniques rely on concurrent parallel lines as the main feature for detecting stairs.
However, traditional computer vision techniques are usually sensitive to changes in lighting conditions and materials, which may not perform well in complex and unfamiliar scenarios. To address this limitation, a deep learning-based approach for stair line segment detection was proposed in~\cite{wang2022deep}.\\
In the field of computer vision, there has been significant research on general-purpose object detection and line segment detection, with a focus on developing algorithms that can run efficiently on mobile devices. Examples of these methods include YOLO~\cite{redmon2016you} and MobileNets~\cite{howard2017mobilenets,sandler2018mobilenetv2} for object detection, and LETR~\cite{xu2021line}, TP-LSD~\cite{huang2020tp}, and M-LSD~\cite{gu2022towards} for line segment detection.
These methods have generally been developed for detecting a wide range of objects including stairs, which are efficient and easy to implement and maintain.
Most previous studies on stair localization for robots have focused on tracked robots, which only require alignment with the stairs rather than position and orientation.
These studies utilized various approaches such as a PID controller~\cite{patil2019deep} and a fuzzy controller~\cite{mihankhah2009autonomous} to align the robot with the stairs. However, these methods may not be sufficient for robots with different types of mobility, because they only provide partial information on the position, orientation, and direction of the stairs.
Fourre et al. proposed an autonomous staircase localization method for tracked robots using depth data~\cite{fourre2020autonomous}.
The method involves detecting the first step of the staircase and using it to localize the rest of the stairs.
The method was tested in a controlled laboratory environment using a VICON system and was also evaluated on industrial stairs, which are more challenging to detect and localize than typical stairs.
Other methods proposed for stair detection and localization include graph-based stair detection and localization (GSDL)~\cite{westfechtel3DGraphBased2016,westfechtel2018}, which can handle various types of staircases, and staircase modelling, detection, and localization (SMDL) method~\cite{perez-yusDetectionModellingStaircases2015,perez-yusStairsDetectionOdometryaided2017}, which is designed for use with a wearable RGB-D camera.
\edit{
In~\cite{zhuATVNavigationComplex2020}, a pointcloud clustering method is used with a LIDAR.
}
These methods rely on depth data for stair detection and localization, which can be susceptible to occlusions or unstructured environments.
In contrast,
vision-based approaches can provide an ROI for the robot to \textit{focus on}, enabling more robust stair localization.
\edit{
In \cite{sanchez-rojasStaircaseDetectionCharacterization2021},
an ROI is extracted using YOLO and a region-growing approach, a classical computer vision technique, is utilized to identify and localize staircases
with a demonstration of an indoor structured staircase.
}

In this study,
an approach for staircase detection and localization
is proposed for autonomous robotic exploration in urban environments
to address the limitations of previous methods.
The method consists of three modules: stair detection, stair line segment detection, and stair localization. The stair detection module uses YOLO~\cite{redmon2016you} to provide a bounding box around the stairs, which serves as an ROI for the stair line segment detection module.
This module uses M-LSD~\cite{gu2022towards}, a deep line segment detector, to extract concurrent parallel lines within the ROI,
and outlier filtering~\cite{fischler1981random} is applied to improve the accuracy of the detected lines.
The stair localization module uses the extracted lines and corresponding depth data to estimate the position, orientation, and direction of the stairs.
The performance of the proposed method is evaluated through hardware experiments and compared with existing methods.
\autoref{fig:banner_fig} shows our autonomous urban exploration system.
The contributions of this study are as follows.
\begin{itemize}
\item \edit{The proposed method can perform robust stair localization against various conditions.}
Unlike existing methods that rely on traditional computer vision techniques or depth data, the proposed method is less sensitive to environmental changes such as lighting conditions and materials.
It also provides information about the position, orientation, and the direction of the stairs, which is useful for a variety of robot types such as quadruped and tracked robots.
\edit{
The performance of the proposed method is compared with existing methods and demonstrated by real-world hardware experiments for various conditions:
i) structured and unstructured staircases, ii) upstairs and downstairs,
iii) lighting conditions, shadows, and dirt, iv) staircases with occlusions by artificial and natural objects,
and v) stairs in indoor and outdoor environments.
The proposed method is also demonstrated with publicly available data collected by a hand-held moving camera and on-board testing during autonomous exploration in an outdoor environment.
}
\item \edit{
Only a single RGB-D camera is required,
and the modular design makes it easy to implement and maintain the proposed framework.
The deep learning networks used in this study are designed for general object detection and line segment detection,
and specific line segment labelling for stair lines is not required unlike an existing study~\cite{wang2022deep}.
}
\end{itemize}
\begin{figure}[!b]
    \centering
    \includegraphics[width=0.60\linewidth]{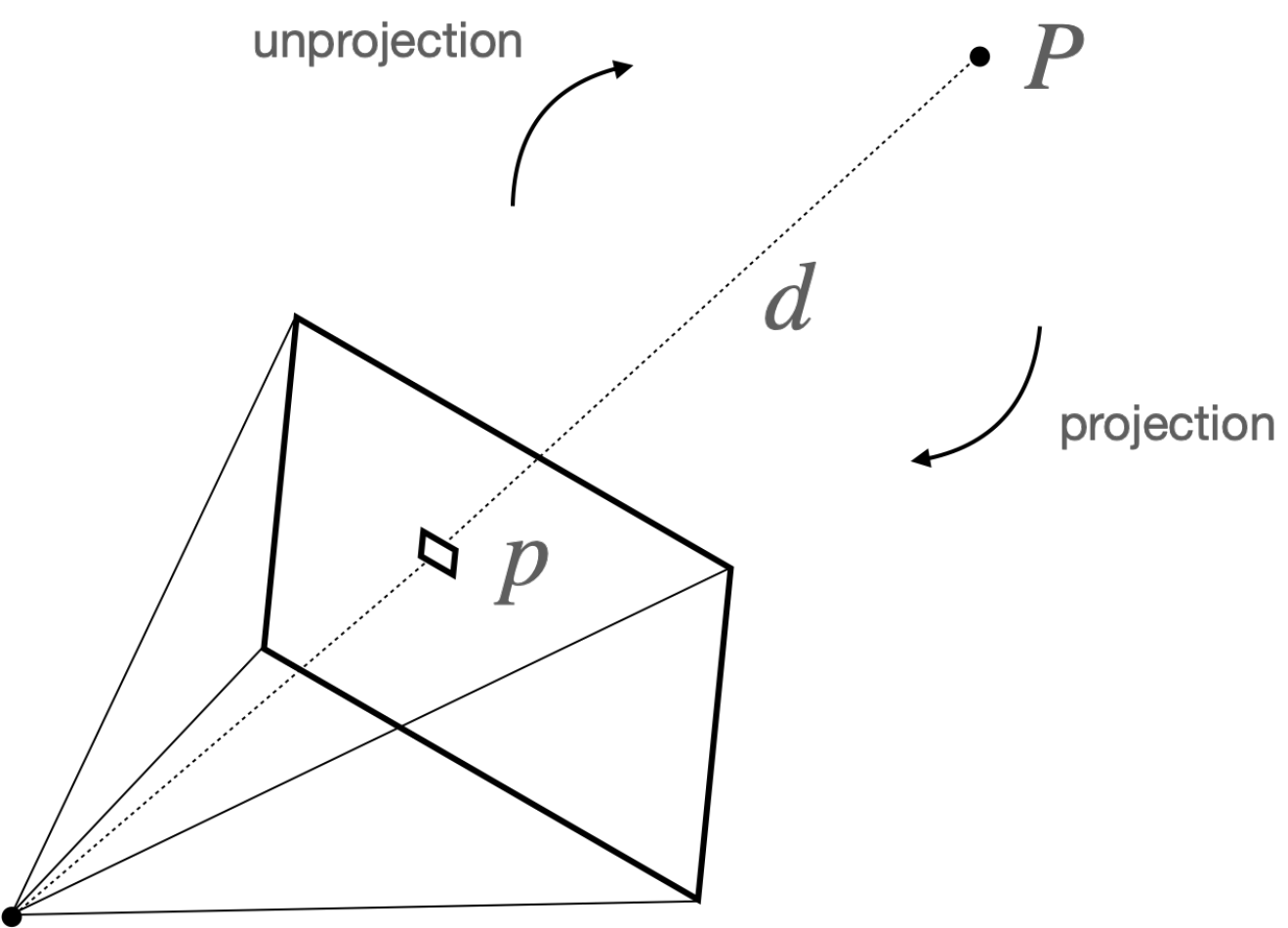} 
    \caption{
    Illustration of the projection of a three-dimensional point
    and unprojection of a pixel.
    The projection and unprojection are performed based on camera specification.
    In this study,
    the unprojection of a pixel is used to generate a point cloud,
    which is mainly used in the stair localization module.
    The projection of a three-dimensional point is used to match the detected line segments and corresponding points in a point cloud.
    }
    \label{fig:proj_unproj}
\end{figure}

\begin{figure}
    \centering
    \includegraphics[width=1.0\linewidth]{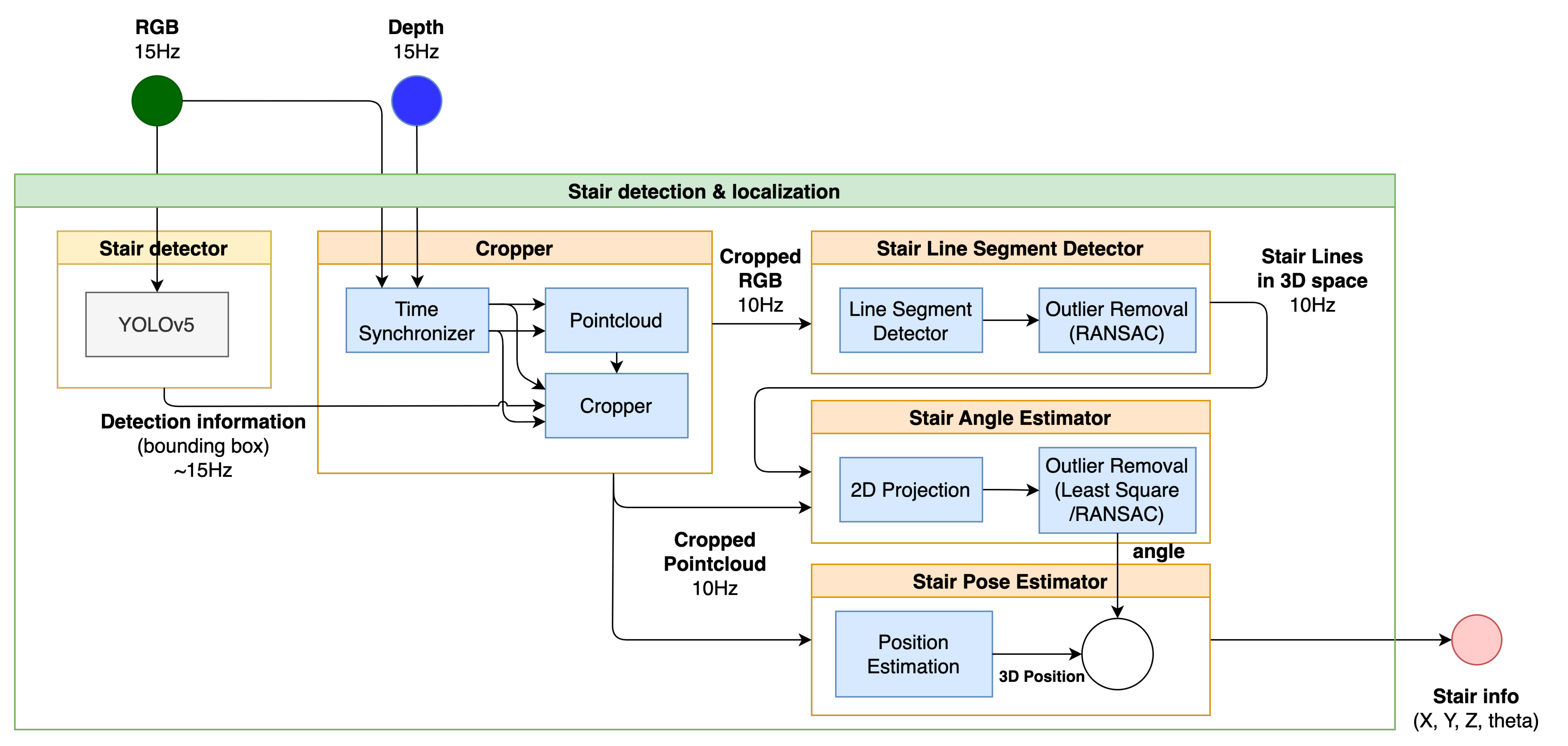}
    \caption{
    Pipeline of the proposed method.
    The proposed method consists of i) stair detection, ii) stair line segment detection, and iii) stair localization (including stair angle estimator and stair pose estimator) modules.
    The RGB color image and point cloud from a single RGB-D camera are used and processed in each module.
    The resulting output of the proposed algorithm includes the position, orientation, and direction (upstairs and downstairs) of the detected staircase.
    }
    \label{fig:pipeline}
\end{figure}

\section{Preliminaries}
Let $\mathbb{N}$ be the set of natural numbers, and $\mathbb{N}_{0} := \mathbb{N} \cup \{0\}$.
For nonnegative integers $i \leq j$, $\mathbb{N}_{i:j}$ denotes the set of natural numbers
equal to or greater than $i$ and equal to or less than $j$,
for example, $\mathbb{N}_{0:10} = \{0, 1, \cdots, 10\}$.
The subscript can be omitted as $\mathbb{N}_{:}$ for brevity.
The set of pixels is denoted by $S = \mathbb{N}_{i_{x}:j_{x}} \times \mathbb{N}_{i_{y}:j_{y}}$ for some nonnegative integers $i_{(\cdot)}$'s and $j_{(\cdot)}$'s.
Let $\mathbb{R}$ be the set of real numbers.
The set of nonnegative and positive real numbers are denoted by $\mathbb{R}_{+}$ and $\mathbb{R}_{++}$, respectively.

\subsection{Object detection and bounding box}
In object detection,
the detection result is provided as a form of bounding box $b = [\underline{x}, \underline{y}, \overline{x}, \overline{y}]^{\intercal} \in S^{2}$,
where $(\underline{x}$, $\underline{y}) \in S$ and $(\overline{x}$, $\overline{y}) \in S$
denote the pixels of the top-left and bottom-right vertices of the bounding box $b$,
respectively.
The object class is omitted here because only stair is considered in this study.

\subsection{Line segment representation}
\label{sec:line_segment_representation}
Line segment representation may influence the network structures,
which affects the inference speed.
Tri-points (TP) representation was proposed in~\cite{huang2020tp} and implemented in~\cite{gu2022towards},
which represents a line segment by its root point and two opposite displacements.
The simple representation enhances inference speed, which makes it possible to perform
line segment detection in real-time.
Note that the TP representation can be expressed as vectorized lines by TP generation~\cite{huang2020tp} as
$p_{\text{start}} =p_{\text{root}} + d_{\text{start}}(p_{\text{root}})$,
and $p_{\text{end}} =p_{\text{root}} + d_{\text{end}}(p_{\text{root}})$,
where $p_{\text{root}}$, $p_{\text{start}}$, $p_{\text{end}} \in S$ denote
the root, start, and end pixels, respectively,
and $d_{\text{start}}(p_{\text{root}})$, $d_{\text{end}}(p_{\text{root}}) \in \mathbb{R}^{2}$
denote the displacements
from the root to start and end points, respectively.
Typically, each pixel $p$ is internally treated as
an element of $\mathbb{R}_{+}^{2}$ and be rounded numerically
to make sure that the pixel $p$ is in $S$.

\subsection{Projection and unprojection of pixels}
Let us consider the standard pinhole camera model~\cite{vasiljevic2020neural}.
The projection and unprojection mappings between pixels and three-dimensional points
can be performed with the camera intrinsic matrix.
Let $P = [X, Y, Z]^{\intercal} \in \mathbb{R}^{3}$ be a three-dimensional point,
where $X$, $Y$, and $Z$ correspond to left, down, and forward, respectively,
with respect to the camera frame.
The projection of $P$ onto the projection plane can be performed as
\begin{equation}
    \label{eq:projection}
    \tilde{p} = \frac{1}{d} K P,
\end{equation}
where $\tilde{p} = [p^{\intercal}, 1]^{\intercal}$ is a pixel in homogeneous coordinates,
$p \in S$ is the corresponding pixel, and $d \in \mathbb{R}_{+}$ is the depth of pixel $p$.
$K \in \mathbb{R}^{3\times3}$ is the camera intrinsics matrix,
whose elements consist of focal lengths and biases to make sure that $p$ is defined in $S$.
The unprojection of $p$ to $P$ can be obtained from \eqref{eq:projection} that
\begin{equation}
    \label{eq:unprojection}
    P = d K^{-1} \tilde{p}.
\end{equation}
Figure \ref{fig:proj_unproj} depicts the illustration
of the projection and unprojection mappings in Eqs. \eqref{eq:projection} and \eqref{eq:unprojection}. 

In this study,
the following hypothesis is assumed,
which is supposed in most studies on stair missions possibly with slight modifications~\cite{munoz2016depth,patil2019deep,lu2005detection}.
\begin{hypothesis}
    \label{hypothesis:parallel_lines}
    Concurrent parallel line segments of stair nosing exist and are majority
    in the region of interest obtained from stair detection module.
\end{hypothesis}

\begin{figure*}
    \centering
    \begin{subfigure}{0.24\textwidth}
        \includegraphics[width=\linewidth]{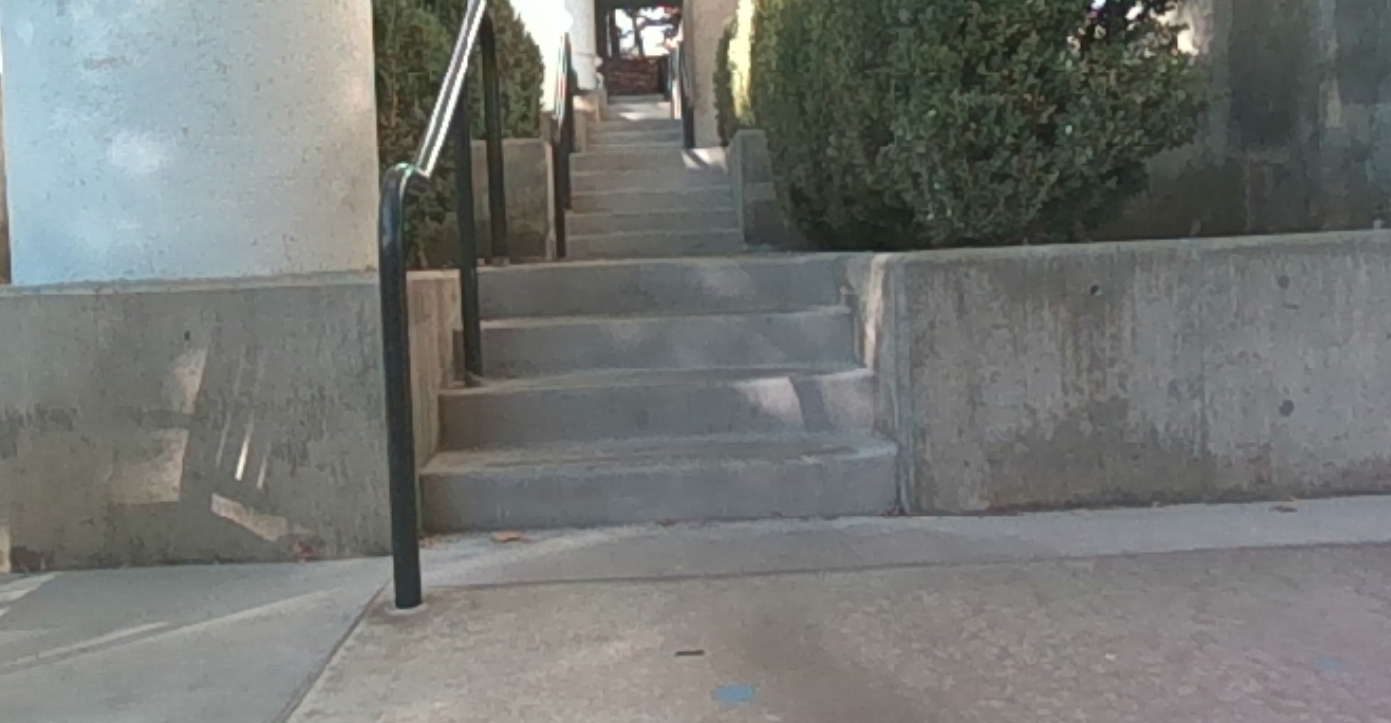}
        \caption{}
        \label{fig:image_example}
    \end{subfigure}%
    \begin{subfigure}{0.24\textwidth}
        \includegraphics[width=\linewidth]{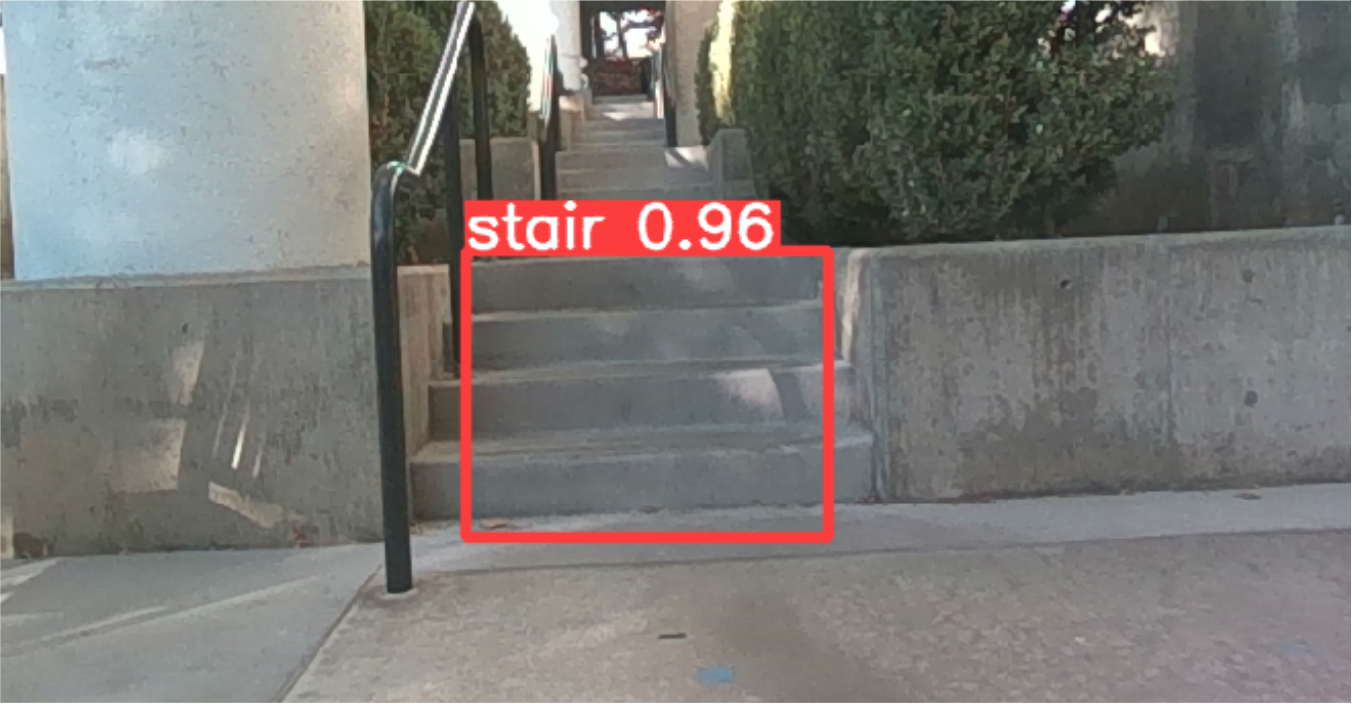}
        \caption{}
        \label{fig:stair_detection}
    \end{subfigure}%
    \begin{subfigure}{0.24\textwidth}
        \includegraphics[width=\linewidth]{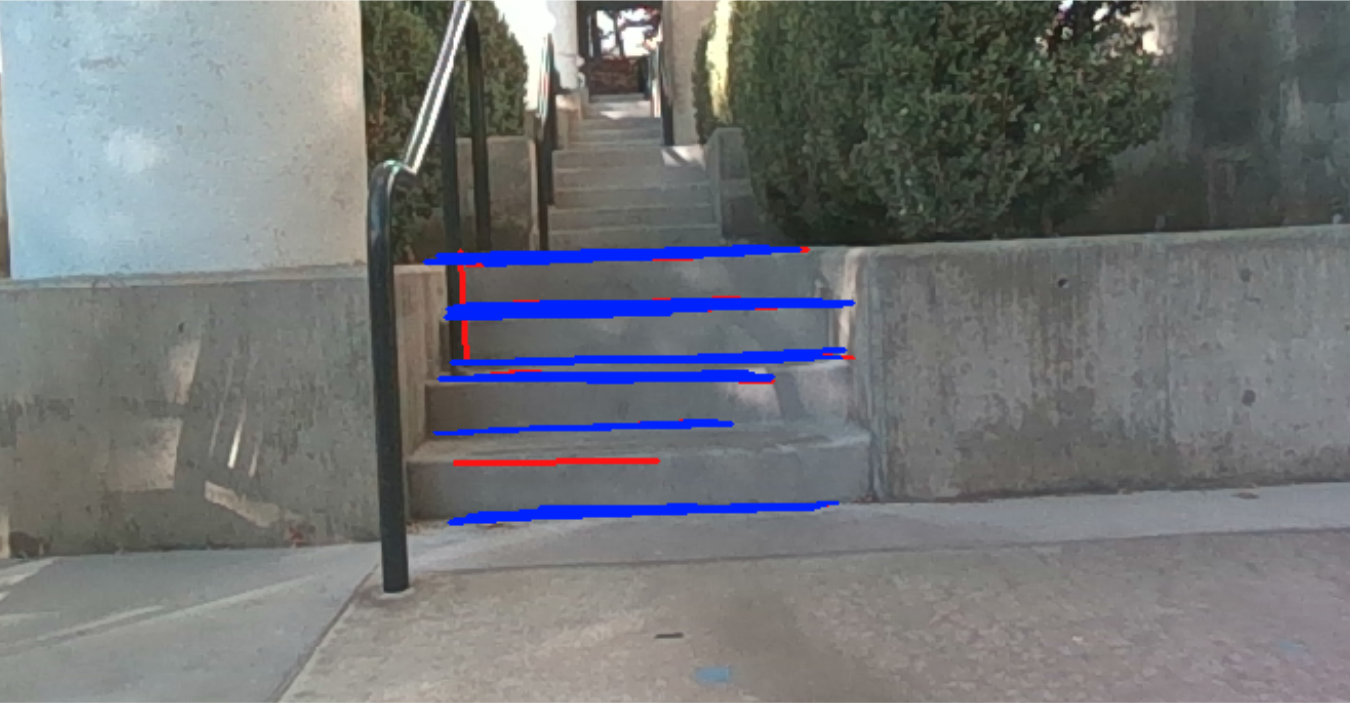}
        \caption{}
        \label{fig:stair_line_segment_detection}
    \end{subfigure}%
    \begin{subfigure}{0.24\textwidth}
        \includegraphics[width=\linewidth, height=0.52\textwidth]{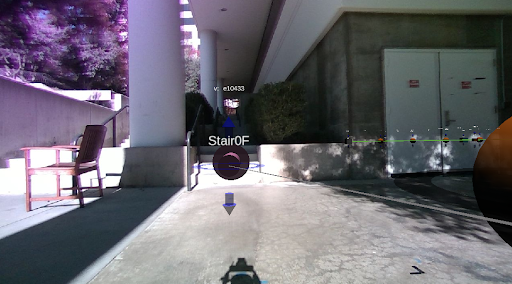}
        \caption{}
        \label{fig:stair_node_generation}
    \end{subfigure}
    \caption{Example outcomes for each component using our proposed method.
    (a) color (RGB) image
    (b) stair detection result with a bounding box
    (c) stair line segment detection result (red lines: detected line segments, blue lines: outlier removed)
    (d) stair localization result (purple arrow: localized staircase in terms of position and orientation)
    }
    \label{fig:example_detection_result}
\end{figure*}

\section{Deep learning based Stair Detection and Localization Algorithm}
\subsection{Overview}
The algorithm proposed in this study consists of multiple modules of
i) stair detection,
ii) stair line segment detection,
and iii) stair localization.
The main idea of the proposed algorithm is to extract the characteristic feature of the stairs, line segments, based on \autoref{hypothesis:parallel_lines}.
Compared to the existing algorithms exploiting the line segments of the stairs,
the proposed algorithm can detect stairs by vision-based deep networks
and perform stair localization for the stair position, orientation, and direction.
Owing to the proposed cascade pipeline,
users can easily obtain insights and understandings from each module,
which makes it easy to improve and maintain the stair localization algorithm.
Also, the stair detection and line segment detection modules do not require
specific networks designed specifically for stairs,
which makes it easy to implement state-of-the-art general-purpose object detection and line segment detection networks.
Note that the proposed method requires only a single RGB-D camera for stair localization.
\autoref{fig:pipeline} summarizes the pipeline of the proposed algorithm,
and \autoref{fig:example_detection_result} illustrates
the stair detection and stair line segment detection results of the proposed method.
The integration with information roadmap (IRM) for autonomous stair climbing is described at the end of this section.

\subsection{Stair detection}
\label{sec:stair_detection}
A stair detection module is designed to provide an ROI for stair line segment detection.
A deep learning algorithm YOLOv5~\cite{Jocher_YOLOv5_by_Ultralytics_2020},
a successor of YOLO~\cite{redmon2016you},
is implemented and trained for stair detection.
The RGB color image stream is received from a single RGB-D camera and provides a bounding box $b$
of the detected staircase.
The streamed color image is cropped using the bounding box $b$.
The cropping can be interpreted as \textit{paying attention} to the stair image.
\autoref{fig:stair_detection} shows an example of
stair detection in this study.

\subsection{Stair line segment detection}
\label{sec:line_segment_detection}
A stair line segment detection module is implemented to extract
the concurrent parallel line segments from a stair image.
The input of the module is
the color image cropped from the stair detection module,
and the output is the detected line segments from the image.
For the line segment detection, M-LSD-tiny network is adopted
with the image input size of 512x512,
which is designed for real-time line segment detection on mobile devices~\cite{gu2022towards}.
M-LSD-tiny is a lighter version of M-LSD based on encoder-decoder architecture~\cite{gu2022towards}.
The encoder network is a part of MobileNetV2~\cite{sandler2018mobilenetv2} and decoder consists of three block types~\cite{gu2022towards}.

\begin{figure}[t!]
    \centering
    \includegraphics[width=0.48\textwidth]{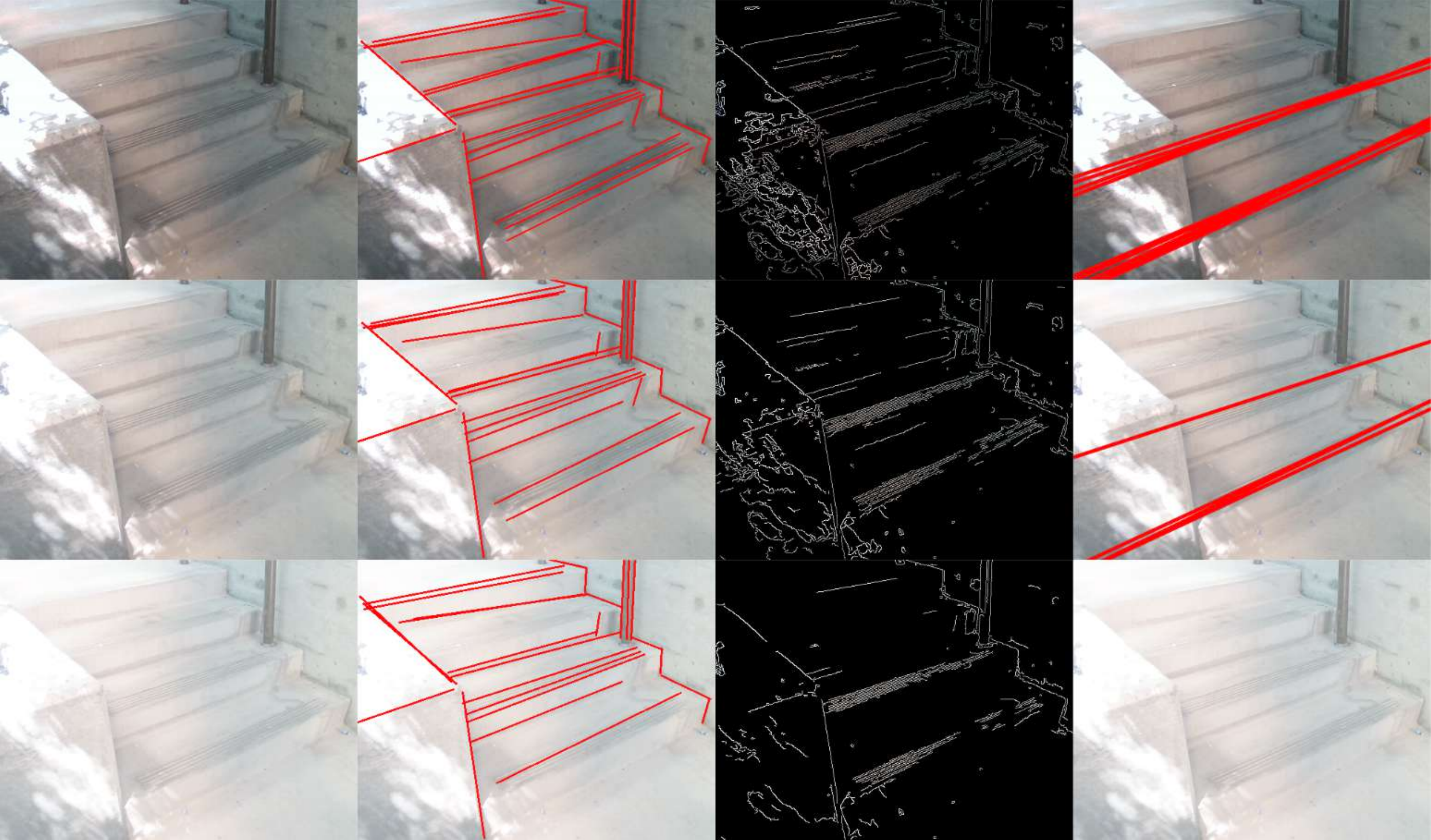}
    \caption{
    \edit{
    The comparison of stair line (segment) detection between the proposed method and a classical CV method (Canny edge detection + Hough transformation).
    Images with different exposures are used for a staircase with indistinct stair lines (the edges of the stair are relatively blunt).
    Each column corresponds to i) the original image, ii) the result of the proposed method, iii) Canny edge detection, and iv) Hough  transformation for line detection.
    The classical CV method could not detect stair lines for the highest exposure case.
    On the other hand, the proposed method detects consistent lines against different exposures.
    }
    }
    \label{fig:classical_vs_deep}
\end{figure}

After line segments are detected using M-LSD-tiny,
outlier lines are picked out to leave only the parallel line segments of the stairs.
RANSAC~\cite{fischler1981random} is used for the outlier removal in terms of the mean squared error (MSE) of line slopes,
based on \autoref{hypothesis:parallel_lines}.
The set of the parallel line segments is denoted by $L$.
Note that each line is expressed by TP representation introduced in \autoref{sec:line_segment_representation}.
The proposed stair line segment detection does not rely only on specific line segment detectors designed for stairs.
\edit{
\autoref{fig:classical_vs_deep} shows that the proposed stair line segment detector is more robust than classical computer vision methods against environmental changes such as lighting conditions.
}

\subsection{Stair localization}
\label{sec:stair_localization}
A stair localization module is designed to estimate the stair pose and direction
from the parallel line segments of stair image.
The module receives a depth image and the parallel line segments detected from the stair line segment detection module,
and provides the stair localization information as output,
which includes stair position, orientation, and direction.

First, a point cloud $\{P_{j} \in \mathbb{R}^{3}\}_{j \in \cup_{l \in L} J_{l}}$ is generated using the depth image,
where $J_{l}$ denotes the set of the indices of the points corresponding to the line segment $l \in L$.
Note that each image pixel $p$ with depth $d$ can be unprojected to a point $P$ in a three-dimensional Euclidean space by Eq. \eqref{eq:unprojection}.
Note also that the intrinsic matrix $K$ in Eq. \eqref{eq:unprojection} is assumed to be known,
which can be obtained from the camera specification.
\autoref{fig:stair_line_segment_detection}) shows an example of line segment detection
in this study.

The stair position $P_{\text{stair}} \in \mathbb{R}^{3}$ is estimated as the average position of the generated point cloud as follows,
\begin{equation}
    \label{eq:stair_position}
    P_{\text{stair}}
    = \frac{1}{\lvert \cup_{l \in L} J_{l} \rvert} \sum_{j \in \cup_{l \in L} J_{l}} P_{j}.
\end{equation}
To estimate the stair orientation, each point cloud $\{P_{j} \in \mathbb{R}^{3} \}_{j \in J_{l}}$ corresponding to a parallel line $l$ is projected onto the ground.
\edit{
For most ground vehicles,
the gravity direction is assumed to be parallel to the $y$-axis of the camera frame.
In other cases, the gravity direction can be obtained by additional sensors and algorithms such as IMU and SLAM.
}
Note that the projection onto the ground, $\pi: \mathbb{R}^{3} \to \mathbb{R}^{2}$ is different from the projection introduced in Eq. \eqref{eq:projection}.
For example, if the camera frame's $Y$-axis is parallel to the local gravity vector,
then the projection onto the ground is $\pi([x, y, z]^{\intercal}) = [x, z]^{\intercal}$.
Then,
each line on the ground, $l_{\text{proj}}$,
is obtained using least squares with the projected points $\{\pi(P_{j}) \}_{j \in J_{l}}$,
which correspond to each parallel line $l$.
The outliers of the lines on the ground are also removed
using RANSAC~\cite{fischler1981random}
in terms of the least-squares error.
This outlier removal is not based on \autoref{hypothesis:parallel_lines}
but to prevent outliers when projecting the points due to the depth noise.
The stair orientation, a unit quaternion $q_{\text{stair}} \in \mathbb{S}^{3}$,
can be obtained from the rotated angle of the stair $\theta_{\text{stair}}$~\cite{lynch2017modern}.
The stair angle $\theta_{\text{stair}}$ is estimated as
\begin{equation}
    \label{eq:stair_angle}
    \theta_{\text{stair}}
    = \frac{1}{\lvert L_{\text{proj}} \rvert} \sum_{l_{\text{proj}} \in L_{\text{proj}}} \theta_{l_{\text{proj}}},
\end{equation}
where $\theta_{l_{\text{proj}}} \in [-\pi, \pi)$ is the angle of projected line $l_{\text{proj}}$.
Also, the stair direction $D_{\text{stair}}$ (up or down) can be determined using the estimated relative height $h_{\text{stair}} \in \mathbb{R}$ of the stair,
which can be obtained from the estimated position $P_{\text{stair}}$.
In this study, the stair direction is determined as
\begin{equation}
    \label{eq:stair_direction}
    D_{\text{stair}} = 
    \begin{cases}
        \text{up}, & \text{if } h_{\text{stair}} > \epsilon \\
        \text{down}, & \text{if } h_{\text{stair}} < -\epsilon \\
        \text{ambiguous}, & \text{otherwise}
    \end{cases}
\end{equation}
where $\epsilon \in \mathbb{R}_{++}$ is a threshold parameter.
Algorithm \autoref{alg:stair_localization} shows the pseudo-code of the proposed stair localization algorithm.

\edit{It is possible to provide different types of localization information, for example, the first step of the staircase by finding the closest stair line segment. In this study, the abstract information of position and orientation is extracted for the integration with information roadmap (IRM), which will be explained in \autoref{sec:integration_with_irm}.}

\begin{algorithm}[!h]
\caption{Proposed stair localization algorithm}
\label{alg:stair_localization}
\begin{algorithmic}[1]
\State \textbf{Initialize: }
$P_{\text{stair}} = [0, 0, 0]^{\intercal}$,
$\theta_{\text{stair}} = 0$,
$D_{\text{stair}} = \text{None}$,
$N_{P}=0$,
$N_{\theta}=0$

\State \textbf{Input: }
RGB Image $I_{\text{RGB}}$,
depth image $I_{\text{depth}}$

\State \textbf{Output: }
Stair position $P_{\text{stair}}$,
angle $\theta_{\text{angle}}$,
direction $D_{\text{stair}}$

\State \textbf{\# Preprocessing}
\State Bounding box $b$ = \texttt{detect\_stair}($I_{\text{RGB}}$) (\autoref{sec:stair_detection})
\State Cropped image $I_{\text{RGB}}^{\text{cropped}}$ = $I_{\text{RGB}}$[$\underline{x}$:$\overline{x}$, $\underline{y}$:$\overline{y}$]
\State Line segments $L$ = \texttt{detect\_segments}($I_{\text{RGB}}^{\text{cropped}}$) (\autoref{sec:line_segment_detection})
\State Point cloud pcd = $\{ P_{j} = \texttt{unproject}(I_{\text{RGB}}[j], I_{\text{depth}}[j])$ $\ \vert \ j \in \cup_{l \in L} J_{l} \}$ (\autoref{eq:unprojection})

\State \textbf{\# Position estimation}
\For{$P_{j} \in $ pcd}
    \State $P_{\text{stair}} \mathrel{+}= P_{j}$ (element-wise)
    \State $N_{P} \mathrel{+}= 1$
\EndFor
\State $P_{\text{stair}} \mathrel{/}= N_{P}$

\State \textbf{\# Angle estimation}
\For{$l \in L$}
    \State $l_{\text{proj}}$ = $\{ \pi(P_{j}) \}_{j \in J_{l}}$
    \If{\texttt{is\_not\_outlier}($l_{\text{proj}}$)}
        \State $\theta_{l_{\text{proj}}}$ = \texttt{obtain\_angle}($l_{\text{proj}}$)
        \State $\theta_{\text{stair}} \mathrel{+}= \theta_{l_{\text{proj}}}$
        \State $N_{\theta} \mathrel{+}= 1$
    \EndIf
\EndFor
\State $\theta_{\text{stair}} \mathrel{/}= N_{\theta}$

\State \textbf{\# Direction determination}
\State $h_{\text{stair}}$ = \texttt{obtain\_height}($P_{\text{stair}}$, $P_{\text{base\_link}}$)
\State $D_{\text{stair}}$ = \texttt{determine\_direction}($h_{\text{stair}}$) (\autoref{eq:stair_direction})

\end{algorithmic}
\end{algorithm}

\subsection{Integration with information roadmap (IRM)}
\label{sec:integration_with_irm}
IRM is a concept for multi-level planning
method, implemented in NeBula and its successors~\cite{agha2021nebula, agha2014firm, kim2021plgrim}.
In this study,
the staircase localization method is integrated with IRM by publishing the pose information
of detected staircases.
\edit{When the proposed staircase localizer detects and localizes a staircase, the staircase information is abstracted as a stair IRM node candidate. If a cluster of stair IRM node candidates is collected with low standard deviation in terms of position and orientation, then a stair IRM node is published, which rejects the registration of new stair IRM nodes near the previously published stair node.}

\section{Experiments}
In this section,
the staircase detection and localization method
proposed in this study is compared with existing methods,
and evaluated in various indoor and outdoor environments.
The performance of the proposed method is demonstrated in a fully autonomous exploration integrated with IRM.
\edit{
Intel RealSense D455 RGB-D camera is used for experiments in this study.
}

\begin{figure}[t!]
    \centering
    \includegraphics[width=1.0\linewidth]{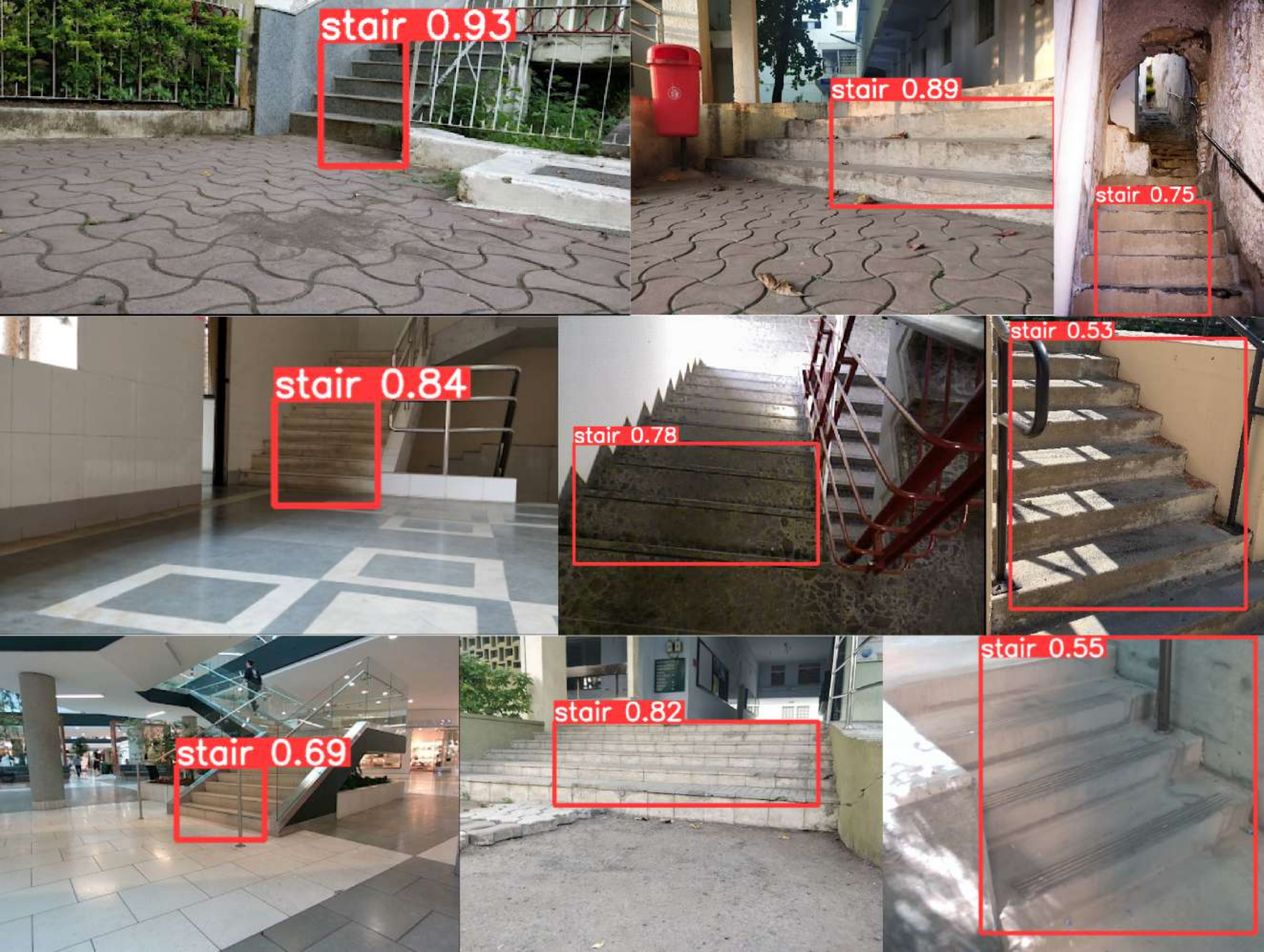}
    \caption{
    \edit{
    Example images in the test dataset used for the stair detector.
    The dataset includes various structured and unstructured upstairs and downstairs in indoor and outdoor environments.
    Most training data are borrowed from Unmesh dataset~\cite{patil2019deep},
    and some staircases are added for the variability of the dataset, obtained manually by the authors.
    }
    }
    \label{fig:dataset}
\end{figure}

\edit{
Total $3\text{,}727$ images are used to train and test the YOLOv5-based stair detector.
The dataset in this study is highly based on Unmesh dataset \cite{patil2019deep}.
Note that the Unmesh dataset contains images mainly from the campus of Visvesvaraya National Institute of Technology, Nagpur and from the Internet,
and the dataset is augmented with horizontal flipping.
To increase the generalization capability of the network,
the authors added uncommon types of staircases to the dataset, including industrial and in-wild staircases.
In addition, the dataset is augmented with many variations for generalization, including different shadows, brightness, random crop, exposure, blur, noise, rotation, and so on.
Test images in the dataset are illustrated in \autoref{fig:dataset}.
For the stair line segment detection,
the pre-trained M-LSD tiny is used~\cite{gu2022towards}.
}

\subsection{Comparison with existing staircase localization methods}
The existing methods compared with the proposed method include
i) Graph-based stair detection and localization (GSDL)~\cite{westfechtel3DGraphBased2016,westfechtel2018},
and ii) Staircase modeling, detection, and localization for a wearable RGB-D camera (SMDL)~\cite{perez-yusDetectionModellingStaircases2015,perez-yusStairsDetectionOdometryaided2017}.
Note that the two existing methods use depth information to detect and localize staircases
in the form of the point cloud.

For comparison,
test data are obtained from a staircase at NASA JPL using an RGB-D camera.
Staircase configurations for comparison are
expressed as the form of \texttt{dist}m-\texttt{location},
the \texttt{dist} denotes the distance from the staircase to the robot (the closest distance on $z$-axis of the camera frame),
and the \texttt{location} means $1$m shift on $x$-axis of the camera frame.

\begin{table*}[!t]
    \centering
    \begin{tabular}{l|c|c|c|c|c}
        \hline
        & $X$ error (m) & $Y$ error (m) & $Z$ error (m) & $\theta$ error (deg) & Remark
        \\   
        \hline
        \hline

        \texttt{JPL 1m-front} (\texttt{out}) & $-0.0011 \pm 0.0177$ & $-0.0111 \pm 0.0331$ & $+0.0359 \pm 0.0793$ & $+2.0450 \pm 0.6007$ & \multirow{7}{*}{Different viewpoints}
        \\        
        \cline{1-5}
        \texttt{JPL 3m-front} (\texttt{out}) & $-0.0042 \pm 0.0355$ & $+0.0528 \pm 0.0485$ & $-0.0185 \pm 0.0875$ & $+1.5472 \pm 1.3337$ &
        \\
        \cline{1-5}
        \texttt{JPL 3m-left} (\texttt{out}) & $+0.0556 \pm 0.0517$ & $+0.2371 \pm 0.0569$ & $-0.1236 \pm 0.1402$ & $+5.0043 \pm 4.3864$ &
        \\               
        \cline{1-5}
        \texttt{JPL 3m-right} (\texttt{out}) & $+0.1413 \pm 0.0729$ & $+0.2222 \pm 0.0961$ & $-0.1345 \pm 0.1815$ & $+3.9765 \pm 3.3354$ &
        \\        
        \cline{1-5}
        \texttt{JPL 5m-front} (\texttt{out}) & $+0.4017 \pm 0.0818$ & $-0.1437 \pm 0.0838$ & $+0.0443 \pm 0.1916$ & $+11.6437 \pm 9.0328$ &
        \\        
        \cline{1-5}
        \texttt{JPL 5m-left} (\texttt{out}) & $+0.2699 \pm 0.0580$ & $+0.0200 \pm 0.0678$ & $-0.1808 \pm 0.1653$ & $+8.2997 \pm 6.2494$ &
        \\        
        \cline{1-5}
        \texttt{JPL 5m-right} (\texttt{out}) & $+0.1010 \pm 0.1210$ & $+0.0620 \pm 0.0937$ & $-0.1258 \pm 0.2119$ & $+13.5457 \pm 10.4206$ &
        \\        
        \hline
        \hline
        
        \texttt{SNU 301} (\texttt{in}) & $+0.4286 \pm 0.0560$ & $-0.0122 \pm 0.0844$ & $+0.1484 \pm 0.1423$ & $+4.9887 \pm 0.4499$ & \multirow{3}{*}{Different stairs}
        \\
        \cline{1-5}
        \texttt{SNU 302} (\texttt{in}) & $+0.0797 \pm 0.0180$ & $-0.1197 \pm 0.0465$ & $-0.0356 \pm 0.0783$ & $+1.5733 \pm 0.7911$ &
        \\    
        \cline{1-5}
        \texttt{SNU 301 to 302} (\texttt{out}) & $-0.0284 \pm 0.0349$ & $+0.0101 \pm 0.0528$ & $+0.1497 \pm 0.1894$ & $+1.2291 \pm 0.8023$ &
        \\
        \hline
        \hline

    \end{tabular}
    \caption{
    \edit{
    Stair localization errors are denoted as $\mu \pm \sigma$ where $\mu$ and $\sigma$ are the mean and standard deviation of the errors, respectively.
    \texttt{in} and \texttt{out} denote indoor and outdoor environments, respectively.
    }
    }
    \label{tab:evaluation}
\end{table*}

\autoref{tab:evaluation} shows the proposed method's errors of
three-dimensional stair position $P_{\text{stair}} = [X, Y, Z]^{\intercal}$
and the stair angle $\theta_{\text{stair}}$
with respect to various configurations.

\begin{figure}[t!]
    \centering
    \includegraphics[width=0.48\textwidth]{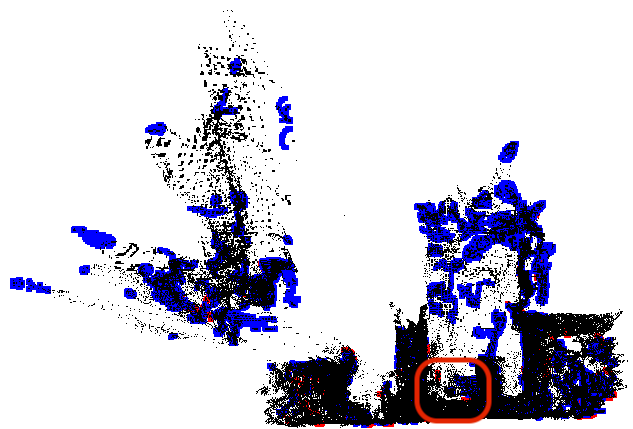}
    \caption{
    GSDL failure case: outdoor environments.
    Staircase is denoted by a red rounded box.
    The point cloud obtained in outdoor environments
    are typically unstructured and staircases take only small portion in the point cloud.
    This makes it hard to detect and localize staircases
    only using point cloud data.
    }
    \label{fig:gsdl_failure_case}
\end{figure}

\begin{figure}[!h]
    \centering
    \begin{subfigure}[b]{0.235\textwidth}
        \centering
        \includegraphics[width=\linewidth]{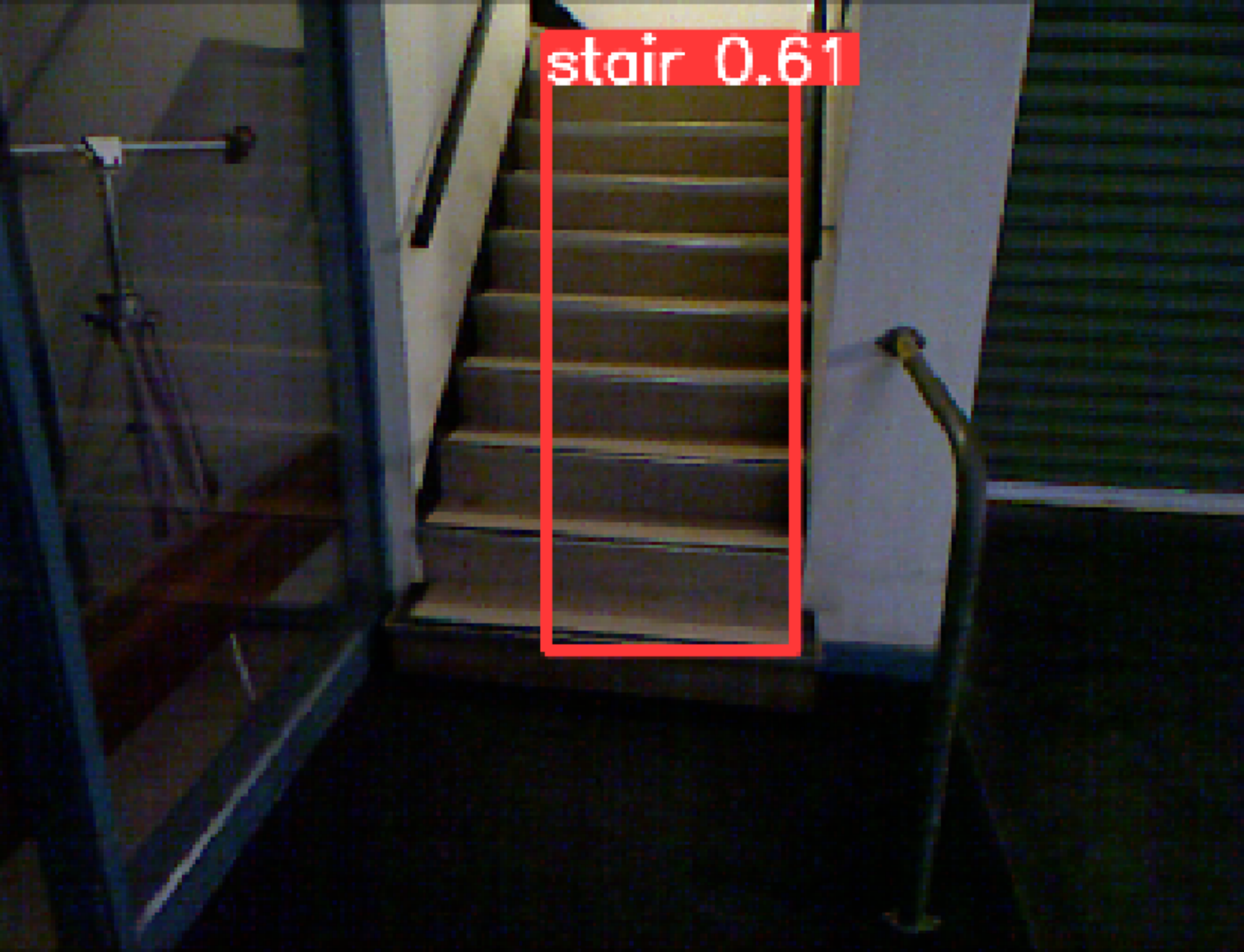}
        \caption{Detected bounding box}
    \end{subfigure}
    \begin{subfigure}[b]{0.24\textwidth}
        \centering
        \includegraphics[width=\linewidth]{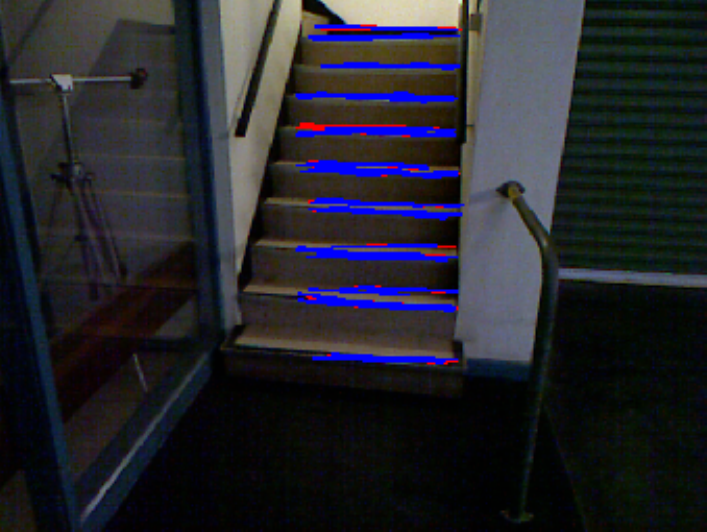}
        \caption{Detected line segments}
    \end{subfigure}
    
     \begin{subfigure}[b]{0.235\textwidth}
        \centering
        \includegraphics[width=\linewidth]{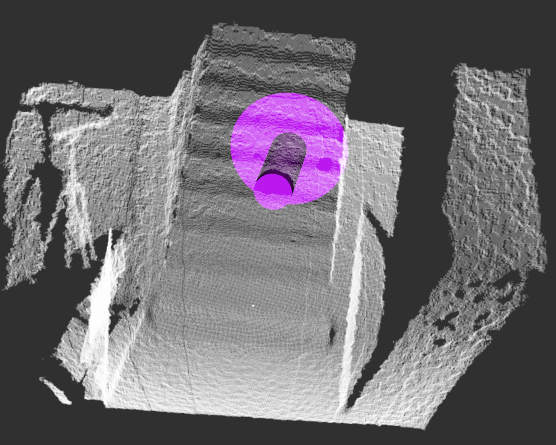}
        \caption{Result of the proposed method}
    \end{subfigure}   
    \begin{subfigure}[b]{0.24\textwidth}
        \centering
        \includegraphics[width=\linewidth]{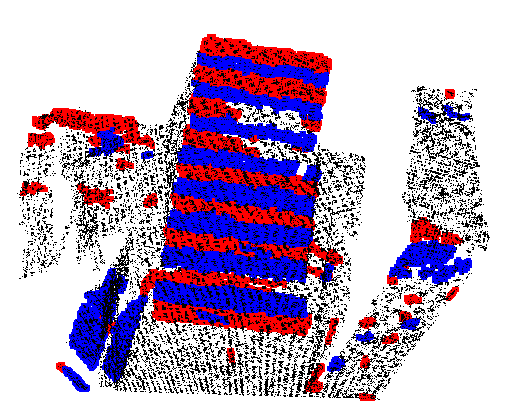}
        \caption{Result of GSDL method}
    \end{subfigure}
    \caption{
    \edit{GSDL success case: indoor environments~\cite{westfechtel3DGraphBased2016,westfechtel2018}.
        Staircases usually take a large portion of the point cloud obtained in indoor environments,
        which makes it relatively easy to detect and localize the staircases using point cloud data.
        (a) The detected bounding box, (b) the detected line segments, (c) the localization result of the proposed method.
        (d) The result of GSDL method.}
    }
    \label{fig:gsdl_success_case}
\end{figure}

\edit{
For $1$m-\texttt{front},
the mean and standard deviation of position and angle errors are all small.
For $3$m-\{\texttt{front},\texttt{left},\texttt{right}\},
the position errors are still small, but the mean angle error is slightly increased up to $5$ deg.
For $5$m-\{\texttt{front},\texttt{left},\texttt{right}\},
the mean position error is increased up to about $0.5$m,
and the mean angle error is increased up to about $13$ deg.
}
Note that the errors tend to be larger as the distance from the staircase
gets farther.
From the hardware experiment result,
it can be concluded that
the proposed method can perform very accurate stair localization up to $5$m in terms of position error
and $3$m in terms of angle error.
It should be pointed out that the two existing methods for comparison
could not detect any staircases or provided wrong detection and localization results
\textbf{for all configurations}.
\edit{
Note that the proposed method also succeeds to localize the test cases of indoor stairs provided by \cite{westfechtel3DGraphBased2016,westfechtel2018}.
An example failure case of GSDL~\cite{westfechtel3DGraphBased2016,westfechtel2018}
is depicted in \autoref{fig:gsdl_failure_case}.
Blue and red patches denote detected stair planes.
As shown in \autoref{fig:gsdl_failure_case},
GSDL regarded random surrounding objects as the detected staircase.
Meanwhile,
\autoref{fig:gsdl_success_case} shows a successful case of GSDL for indoor staircase detection and localization using data provided from~\cite{westfechtel3DGraphBased2016,westfechtel2018}.
Note that the proposed method also succeeds to localize the indoor staircase.
It can be deduced from the success and failure cases of GSDL that
the stair detection and localization only using point cloud may fail and show poor performance
particularly for outdoor staircases
because point clouds in outdoor environments would have unstructured depth information
within various depth ranges,
which makes it hard to detect staircases.
On the other hand, the proposed method can pay attention to the ROI by using RGB-based deep stair detection,
which makes it possible to detect and localize the staircase in outdoor environments.
}
\begin{figure*}[!t]
    \centering
    \includegraphics[width=1.0\linewidth]{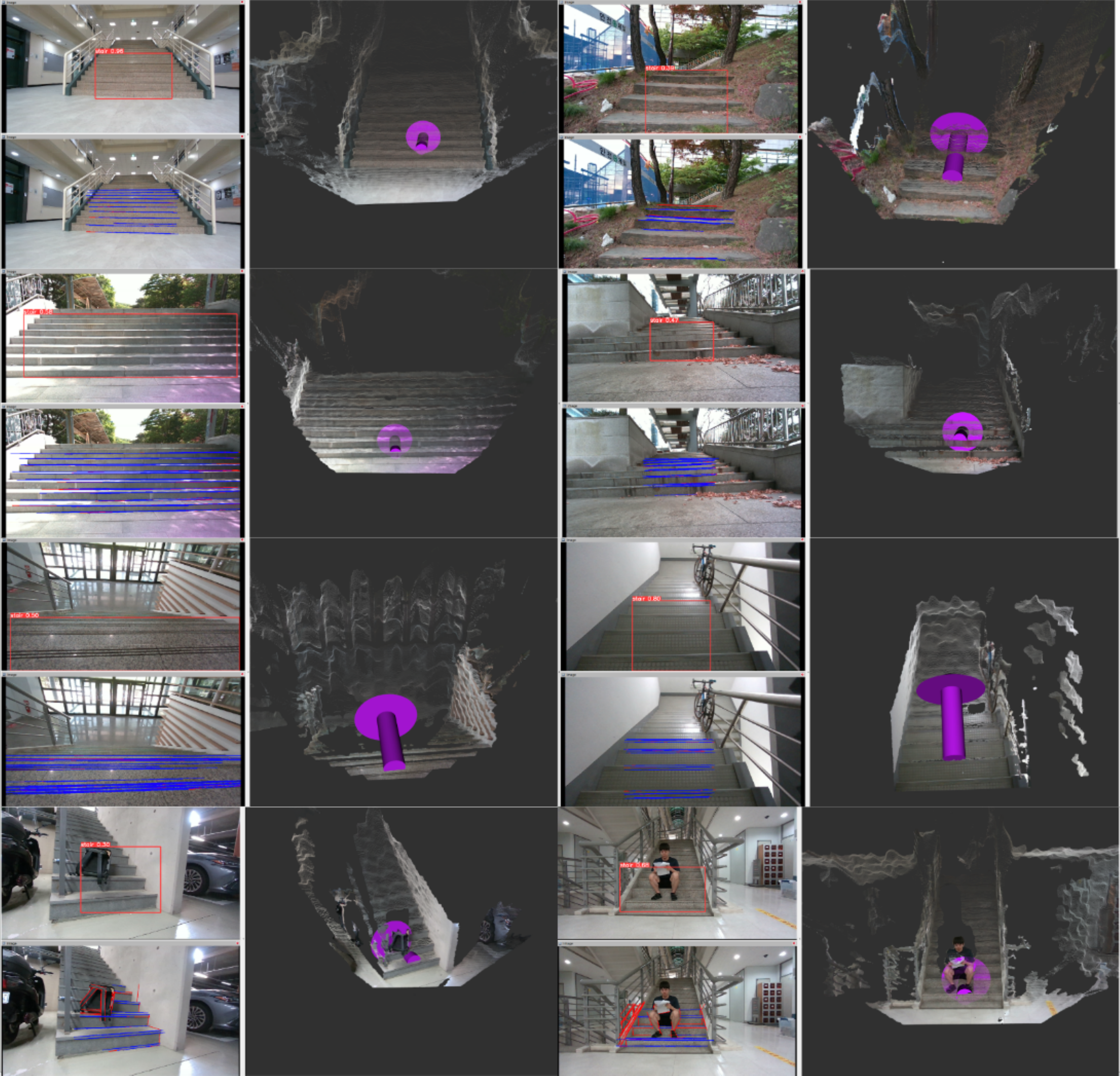} 
    \caption{
    \edit{Qualitative results of the proposed method for staircases with various conditions
    including i) structured and unstructured stairs, ii) shadow and dirt, iii) upstair and downstair, iv) stairs with occlusion by artifical and natural objects.
    Purple arrows indicate the localization results including the localized stair's position and orientation.}
    }
    \label{fig:evaluation}
\end{figure*}

\begin{figure}
    \centering
    \includegraphics[width=0.85\linewidth]{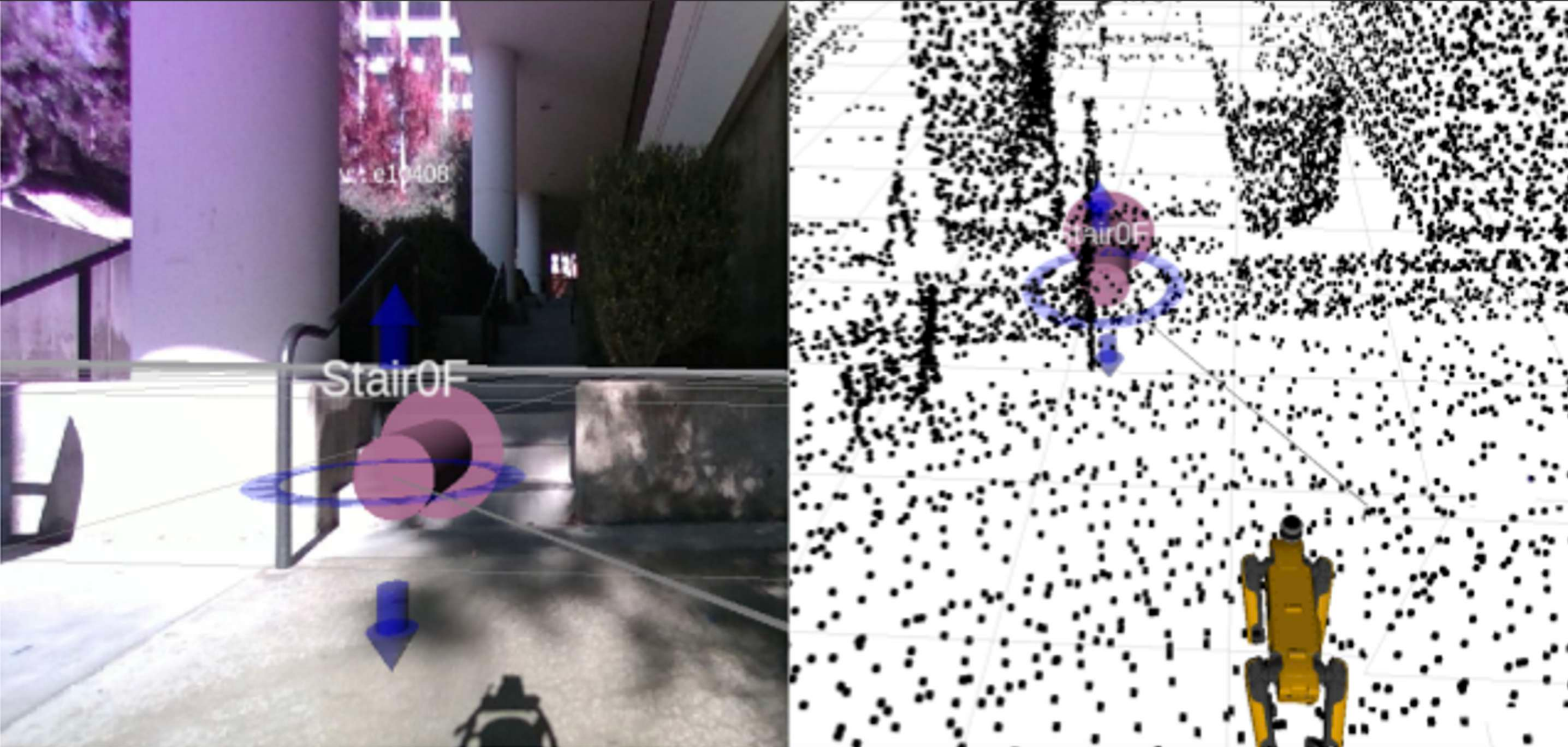} 
    \caption{
    \edit{
    Visualization of the demo scenario.
    (left) The camera view and (right) the top view reconstructed with overlayed stair IRM node.
    Video of the demo can be found in supplementary material.
    }
    }
    \label{fig:stair_node}
\end{figure}

\subsection{Evaluation of stair localization in various indoor and outdoor environments}
In addition to the staircase at NASA JPL,
the proposed stair localization method is evaluated with staircases (for 3m-\texttt{front} configuration) at Seoul National University (SNU) considering indoor and outdoor environments:
i) \texttt{SNU 301}: staircase in SNU building 301 (indoor),
ii) \texttt{SNU 302}: staircase in SNU building 302 (indoor),
and iii) \texttt{SNU 301 to 302}: staircases connecting building 301 to building 302 (outdoor).
\edit{
Note that some images of the tested SNU staircases are included in the qualitative result in \autoref{fig:evaluation}.
}
\autoref{tab:evaluation} shows the stair localization results in terms of the mean and standard deviation of localization errors.
\edit{
The mean position errors and the mean angle error are less than $0.15$m and $5$ deg, respectively, for all tested stairs.
}
This evaluation supports that the proposed method can perform stair localization for staircases in various indoor and outdoor environments.
\edit{
\autoref{fig:evaluation} shows the corresponding qualitative results of this evaluation considering various different conditions.
It should be pointed out that the stair line segment detector detects the line segments of stairs divided by occlusions and effectively rejects the line segments of artificial and natural objects, such as a backpack and a person, by outlier removal.
}
\edit{
Also, the proposed method is evaluated with Bonn stair dataset, a publicly available dataset~\cite{elmasryStairsDetectionModelling2013}.
The data in Bonn stair dataset were collected with a moving hand-held camera at the LBH building of the University of Bonn.
The staircases are indoor and structured but blurred due to the moving camera.
We categorized $100$ upstairs and $30$ downstairs in the dataset as test data, which are not used for training the proposed modules.
The proposed method successfully localized all upstairs ($100 \%$) and $27$ downstairs ($90 \%$) among the test staircases.
The performance was degraded for downstairs because of the relatively narrow field-of-view.
}

\subsection{Demo: Stair localization in autonomous exploration}
Staircase localization is demonstrated with IRM during autonomous robotic exploration.
The proposed method publishes a stair node to IRM if any staircase is detected and localized in exploration.
Note that no active source seeking behavior is implemented to improve staircase localization performance.
\autoref{fig:stair_node} shows
\edit{an example stair IRM node} generated by the proposed method for an upstair at NASA JPL.

\subsection{\edit{Discussion}}
\edit{The proposed method could detect and localize staircases in various environments including indoor and outdoor, structured and unstructured ones with different lighting conditions and occlusions owing to the generalization power of deep learning with various data.
However, this implies that the proposed method inherits the limitation of deep learning techniques:
the proposed method may fail to localize a staircase, which is very different from the training dataset.
This issue is typically referred to as the generalization and domain adaptation of deep learning methods.
Also, it may be hard to extract the correct stair position and orientation for staircases highly occluded by obstacles.
Nonetheless, the proposed cascade design has several advantages. First, each module can be replaced by improved methods, for example, the stair detector may be replaced with a domain-adaptable detector.
Second, even if the stair detector detects a false positive, it is likely rejected by other modules of stair line segment detectors and localizers because the modules qualify given candidate staircases by e.g. RANSAC.
}
\section{Conclusion}

In this paper, we developed a novel approach for localizing stairs for the autonomous exploration of urban environments by a legged robot. Our methodology consists of three modules, including stair detection, line segment extraction, and stair localization.
The stair detection module utilizes a deep learning-based object detection algorithm to generate a region of interest from which the line segment features are extracted using a deep line segment extraction method.
The extracted line segments are then used to determine the staircase's position and orientation.
\edit{
This approach is cost-effective and suitable for various robotic platforms because it can perform stair localization accurately only with a single RGB-D camera.
}
In addition, each module of the proposed pipeline can easily be updated with the latest deep-learning detection techniques.
\edit{
Our real-world experiments have demonstrated that the proposed method can detect and localize different staircases in various indoor and outdoor environments with different conditions such as occlusions by physical objects and particularly surpasses existing methods in complex outdoor urban environments.
}

\section*{ACKNOWLEDGMENT}
The work is partially supported by the Jet Propulsion Laboratory, California Institute of Technology, under a contract with the National Aeronautics and Space Administration (80NM0018D0004).

This work is partially supported by the National Research Foundation of Korea (NRF) grant funded by the Korea government (MSIT) (No. 2019R1A2C2083946).

\bibliographystyle{IEEEtran} 
\bibliography{references.bib}

\end{document}